\documentclass{article}

\usepackage[preprint]{neurips_2026}
\usepackage[T1]{fontenc}
\usepackage{hyperref}
\usepackage{url}
\usepackage{booktabs}
\usepackage{amsmath,amssymb,amsthm}
\usepackage{mathtools}
\usepackage{bm}
\usepackage{graphicx}
\usepackage{float}
\usepackage[table]{xcolor}
\usepackage{algorithm}
\usepackage{algpseudocode}
\usepackage{microtype}
\usepackage{cleveref}
\usepackage{natbib}
\usepackage{multirow}
\usepackage{subcaption}

\newtheorem{theorem}{Theorem}[section]

\newtheorem{lemma}[theorem]{Lemma}
\newtheorem{corollary}[theorem]{Corollary}

\newcommand{\btheta}{{\bm{\theta}}}
\newcommand{\bg}{{\bm{g}}}

\newcommand{\cL}{\mathcal{L}}

\newcommand{\cW}{\mathcal{W}}
\newcommand{\cN}{\mathcal{N}}
\newcommand{\cO}{\mathcal{O}}

\newcommand{\RR}{\mathbb{R}}
\newcommand{\EE}{\mathbb{E}}

\newcommand{\sgop}{\mathrm{sg}}
\newcommand{\diag}{\mathrm{diag}}
\newcommand{\tr}{\mathrm{tr}}
\newcommand{\norm}[1]{\left\|#1\right\|}

\DeclareMathOperator{\proj}{proj}

\title{AEGIS: Anchor-Enforced Gradient Isolation for\\Knowledge-Preserving Vision-Language-Action Fine-Tuning}

\author{%
  Guransh Singh\\
  Independent Researcher\\
  \texttt{guransh766@gmail.com}
}

\begin{document}
\maketitle

\begin{abstract}
Fine-tuning pre-trained Vision-Language Models (VLMs) for robotic manipulation introduces a fundamental stability-plasticity trade-off: continuous flow-matching action experts backpropagate concentrated, low-rank regression gradients into transformer backbones trained on high-dimensional cross-entropy objectives. This \emph{cross-modal gradient asymmetry} rapidly degrades pre-trained visual reasoning and question-answering capabilities. Existing solutions either disconnect continuous gradient flow via stop-gradients (relying on discrete token proxies) or constrain parameter updates via low-rank adaptation (LoRA), which restricts update rank but remains directionally blind to semantic feature corruption. Both approaches typically depend on mixed-batch VQA co-training, doubling training compute. We introduce AEGIS (\emph{Anchor-Enforced Gradient Isolation System}), a buffer-free, layer-wise orthogonal gradient projection framework that enables continuous flow-matching fine-tuning while isolating pre-trained representations from destructive parameter updates. Prior to training, AEGIS estimates per-layer Gaussian activation statistics from pre-training data to establish a static reference anchor. During fine-tuning, a closed-form Wasserstein-$2$ transport penalty generates an anchor-restoration gradient through the active computation graph. A sequential dual-backward pass then applies layer-wise Gram--Schmidt orthogonalization, projecting task gradients onto the orthogonal complement of the restoration vector when directional conflict occurs. We establish an exact energy preservation bound for layer-wise orthogonal projection, showing that AEGIS sheds only $0.62\%$ of gradient energy empirically while halting cumulative feature drift. On PaliGemma2-3B fine-tuned on the LIBERO manipulation benchmark, AEGIS fully preserves pre-trained Visual Question Answering performance and baseline holdout loss while matching the continuous action convergence rate of unconstrained fine-tuning, without requiring replay buffers, teacher models, or co-training data.
\end{abstract}

\section{Introduction}
\label{sec:intro}

Vision-Language-Action (VLA) models adapt web-scale pre-trained Vision-Language Models (VLMs) for robotic manipulation by fine-tuning on action-labelled trajectories \citep{brohan2023rt2, black2024pi0} to transfer multimodal reasoning to physical control.
However, adapting foundation models for robotic control introduces a manifestation of the classical \emph{stability-plasticity dilemma} \citep{mccloskey1989catastrophic, mermillod2013stability}.
While the VLM backbone contains billions of parameters optimized over multimodal cross-entropy pre-training, fine-tuning this backbone on physical action trajectories introduces a low-dimensional regression objective that can rapidly overwrite the pre-trained visual and linguistic representations.

\paragraph{The cross-modal gradient asymmetry.}
A VLM pre-trained on multimodal cross-entropy distributes its representational energy across a broad, high-dimensional singular spectrum to differentiate among hundreds of thousands of token classes.
In contrast, continuous flow-matching action experts \citep{lipman2023flow, black2024pi0} generate high-magnitude Mean Squared Error (MSE) gradients targeting a low-dimensional physical subspace (e.g., $7$-DoF end-effector velocities).
When these concentrated MSE gradients backpropagate unconstrained into the VLM backbone, standard first-order optimizers (SGD, AdamW) project the top-heavy physical updates across the entire parameter space.
This process, which we formalize as \emph{cross-modal gradient asymmetry}, systematically corrupts higher-order semantic dimensions.
In our experiments, naive full-parameter fine-tuning causes severe degradation in Visual Question Answering (VQA) holdout performance within the first $300$ steps (\Cref{fig:trajectories}a), doubling the holdout loss and impairing pre-trained visual reasoning.

\begin{figure}[t]
\centering
\includegraphics[width=0.46\textwidth]{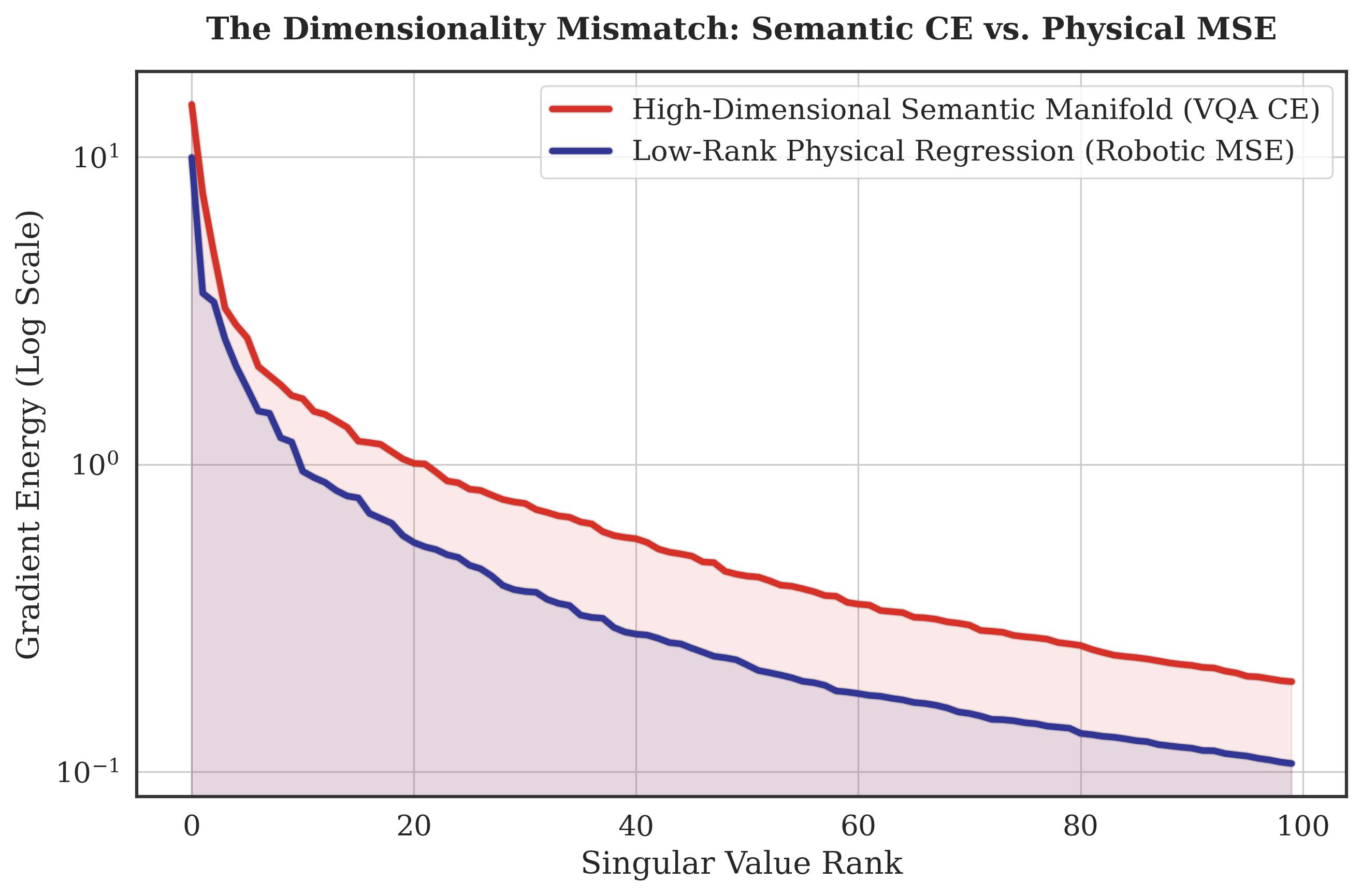}
\caption{\textbf{Empirical evidence of cross-modal gradient asymmetry.}
Singular-value spectrum of real gradients extracted from an intermediate feed-forward down-projection layer (Layer 16) of PaliGemma2-3B.
The {\color{red}VQA cross-entropy gradient} (red) maintains substantial energy across high-order singular dimensions, spanning the broad semantic manifold required for $257{,}000$-class vocabulary prediction.
The {\color{blue}robotic flow-matching MSE gradient} (blue) exhibits steep spectral collapse, concentrating $>90\%$ of its energy in the top $\sim 20$ singular dimensions corresponding to $7$-DoF physical regression.}
\label{fig:spectral}
\end{figure}

\paragraph{The failure of existing defenses and the co-training tax.}
Current VLA systems employ two primary workarounds to mitigate representational collapse.
First, \emph{stop-gradient routing} ($\sgop$) \citep{driess2025knowledge} severs continuous flow-matching gradients at the VLM interface ($\nabla_\btheta \cL_\text{FM} = \mathbf{0}$), delegating VLM adaptation to discrete action tokens (FAST \citep{pertsch2025fast}) trained autoregressively. While stop-gradient protects representations, it eliminates continuous supervision from motor dynamics.
Second, \emph{low-rank adaptation} (LoRA) \citep{hu2022lora, hancock2025vlm2vla} constrains parameter updates to a low-rank bottleneck ($\Delta W = BA$). While LoRA restricts update \emph{rank}, it remains directionally blind: unaligned MSE gradients project directly into the adapter subspace, causing steady VQA erosion (\Cref{fig:trajectories}a).
Consequently, production VLAs universally rely on \emph{mixed-batch VQA co-training}, interleaving up to $50\%$ web VQA data into robotic batches \citep{driess2025kiva, black2024pi0}, which doubles training FLOPs and complicates data pipelines.

\paragraph{AEGIS: Buffer-free orthogonal gradient projection on the Wasserstein manifold.}
To resolve cross-modal gradient asymmetry at its geometric source without co-training data or replay buffers, we propose AEGIS (\emph{Anchor-Enforced Gradient Isolation System}).
AEGIS operates directly on the backward pass via three components: (i) a \emph{static Wasserstein anchor} (\Cref{sec:anchor}) pre-computing per-layer Gaussian activation statistics $(\bm{\mu}^0_\ell, \bm{\sigma}^{0\,2}_\ell)$ from masked VQA forward passes across all 26 language transformer layers and the multi-modal projector; (ii) a closed-form \emph{Wasserstein-$2$ transport penalty} (\Cref{sec:ot-loss}) generating an exact anchor-restoration gradient $\bg_\text{ot}$ through the active graph via the Bures metric; and (iii) a \emph{layer-wise orthogonal gradient projection} (\Cref{sec:ogp}) extracting task and anchor gradients via a sequential dual-backward pass to subtract destructive components ($\langle \bg_\text{task}^\ell, \bg_\text{ot}^\ell \rangle < 0$).
Crucially, AEGIS operates at layer-wise granularity, avoiding cross-layer gradient cancellation and preserving attention head synchronization, while maintaining $99.38\%$ of constructive gradient energy and stabilizing underlying visual-semantic representations.

\paragraph{Contributions.}
(i) We formalize \emph{cross-modal gradient asymmetry} in VLAs, validating the spectral dimensionality mismatch between continuous regression and cross-entropy pre-training (\Cref{sec:asymmetry}).
(ii) We introduce \emph{AEGIS}, a buffer-free framework using static Wasserstein-$2$ anchors and layer-wise Gram--Schmidt projection to shield pre-trained representations during continuous fine-tuning (\Cref{sec:method}).
(iii) We prove that layer-wise Gram--Schmidt projection preserves $(1 - \cos^2\theta_\ell)$ of task energy, shedding only $0.62\%$ of gradient energy empirically (\Cref{sec:ogp}).
(iv) Fine-tuning PaliGemma2-3B on the LIBERO manipulation benchmark, AEGIS achieves complete VQA preservation ($60.23\%$ vs.\ $60.15\%$ baseline on OK-VQA) and baseline CE holdout loss ($0.374$) while matching continuous action convergence without co-training data (\Cref{sec:results}).

\section{Related Work}
\label{sec:related}

\paragraph{Vision-Language-Action Models and Adaptation Paradigms.}
Vision-Language-Action models bridge high-level semantic reasoning and low-level physical control. Early systems discretized continuous robot actions into autoregressive tokens \citep{brohan2022rt1, brohan2023rt2, kim2024openvla}. More recently, $\pi_0$ \citep{black2024pi0} and $\pi_{0.5}$ \citep{driess2025kiva} and knowledge insulation (KIVA) \citep{driess2025knowledge} integrated continuous flow-matching action experts \citep{lipman2023flow}. To prevent flow gradients from corrupting VLM representations, KIVA introduced a strict stop-gradient operator on the flow expert, delegating VLM adaptation to discrete FAST tokens \citep{pertsch2025fast}. Concurrently, VLM2VLA \citep{hancock2025vlm2vla} proposed LoRA fine-tuning \citep{hu2022lora}. As we show in \Cref{sec:results}, LoRA constrains parameter capacity but remains directionally blind, allowing destructive gradient components to overwrite semantic representations. AEGIS resolves this dilemma by operating on gradient \emph{geometry}, enabling direct continuous fine-tuning without stop-gradients or discrete tokenization proxies.

\begin{table}[t]
\centering
\small
\caption{\textbf{Taxonomy of continual learning and gradient surgery paradigms.}
AEGIS provides geometric gradient isolation on curved representation manifolds without requiring replay buffers, live multi-task data streaming, or live teacher forward passes.}
\label{tab:cl-taxonomy}
\setlength{\tabcolsep}{3.0pt}
\begin{tabular}{@{}lccccc@{}}
\toprule
\textbf{Method} & \textbf{Mechanism} & \textbf{Buffer?} & \textbf{2nd Stream?} & \textbf{Scope} & \textbf{Geometry} \\
\midrule
\shortstack[l]{EWC\\\scriptsize\citep{kirkpatrick2017ewc}} & Loss penalty & None & No & Diagonal & Fisher \\
\addlinespace[2pt]
\shortstack[l]{A-GEM\\\scriptsize\citep{chaudhry2019agem}} & Grad inequality & Data & Yes ($2\times$) & Global & Euclidean \\
\addlinespace[2pt]
\shortstack[l]{OGD\\\scriptsize\citep{farajtabar2020ogd}} & Orthogonal proj & Gradients & No & Global & Euclidean \\
\addlinespace[2pt]
\shortstack[l]{GPM\\\scriptsize\citep{saha2021gpm}} & Subspace proj & Activations & No & Per-Tensor & Euclidean \\
\addlinespace[2pt]
\shortstack[l]{PCGrad\\\scriptsize\citep{yu2020pcgrad}} & Multi-task surgery & None & Yes ($2\times$) & Global & Euclidean \\
\midrule
\rowcolor{blue!6}
\textbf{AEGIS (Ours)} & \textbf{Manifold OGP} & \textbf{None (Buffer-free)} & \textbf{No ($1\times$ data)} & \textbf{Layer-Wise} & \textbf{Wasserstein-$\mathbf{2}$} \\
\bottomrule
\end{tabular}
\end{table}

\paragraph{Continual Learning, Gradient Surgery, and Optimal Transport.}
Continual learning seeks to preserve past knowledge during adaptation \citep{de2021continual}. Quadratic penalties \citep{kirkpatrick2017ewc, zenke2017synaptic} are readily overpowered by high-magnitude task gradients in early training.
Orthogonal gradient methods project task gradients away from protected subspaces using stored gradients \citep{farajtabar2020ogd}, live replay buffers \citep{chaudhry2019agem}, activation bases \citep{saha2021gpm}, or multi-task surgery \citep{yu2020pcgrad}. As shown in \Cref{tab:cl-taxonomy}, prior projection methods require explicit replay buffers or concurrent multi-task data streams ($2\times$ compute). AEGIS introduces a distinct paradigm: deriving projection references from a static Wasserstein-$2$ manifold anchor \citep{dowson1982frechet, villani2009optimal, peyre2019computational}, requiring zero replay buffers and no secondary data stream.

\section{Problem Formalisation}
\label{sec:problem}

\paragraph{System setup and action formulation.}
Let $f_\btheta: \mathcal{X} \to \mathcal{H}$ denote a Vision-Language Model pre-trained on multimodal cross-entropy $\mathcal{D}_\text{VLM}$. We fine-tune $f_\btheta$ on robotic trajectories $\mathcal{D}_\text{action} = \{(\bm{x}_i, \bm{a}_i)\}_{i=1}^N$, where $\bm{x} = (\bm{I}_\text{agent}, \bm{I}_\text{wrist}, \bm{p}, \bm{c})$ contains multi-view RGB, proprioception $\bm{p}$, and language $\bm{c}$. The target action $\bm{a} \in \RR^{H \times D_a}$ spans horizon $H=50$ and dimension $D_a=7$.
A continuous flow-matching action expert $\bm{v}_{\bm{\phi}}$ \citep{lipman2023flow, black2024pi0} cross-attends to VLM hidden states $\bm{h}_\btheta = f_\btheta(\bm{x})$ under the flow-matching objective:
\begin{equation}
  \cL_\text{FM}(\btheta, \bm{\phi})
  = \EE_{t, \bm{\epsilon}, (\bm{x}, \bm{a}_1)}
  \left[\norm{\bm{v}_{\bm{\phi}}\!\left(\bm{a}_t; \bm{h}_\btheta, t\right)
  - (\bm{a}_1 - \bm{\epsilon})}^2_2\right],
  \label{eq:fm-loss}
\end{equation}
where $t \sim \text{Beta}(1.5, 1.0)$, $\bm{\epsilon} \sim \mathcal{N}(\mathbf{0}, \bm{I})$, and $\bm{a}_t = t\bm{a}_1 + (1-t)\bm{\epsilon}$ is the conditional flow path.

\paragraph{Cross-modal gradient asymmetry.}
\label{sec:asymmetry}
Let $\nabla_{\btheta_\ell} \cL \in \RR^{d_\text{out} \times d_\text{in}}$ denote the matricized gradient tensor for a linear projection weight matrix at layer $\ell$ under loss $\cL$. With ordered singular values $\sigma_1 \geq \sigma_2 \geq \cdots \geq \sigma_r > 0$ from SVD $\nabla_{\btheta_\ell} \cL = \bm{U} \bm{\Sigma} \bm{V}^\top$, the \emph{spectral concentration ratio} $\kappa_k(\cL) \in (0, 1]$ is:
\begin{equation}
  \kappa_k(\cL) = \frac{\sum_{i=1}^{k} \sigma_i^2}{\sum_{i=1}^{r} \sigma_i^2} = \frac{\sum_{i=1}^{k} \sigma_i^2}{\|\nabla_{\btheta_\ell} \cL\|_F^2}.
  \label{eq:spectral-ratio}
\end{equation}
\emph{Cross-modal gradient asymmetry} occurs when a model pre-trained on an objective with broad spectral energy ($\kappa_k(\cL_\text{CE}) \ll 1$ for small $k$) is fine-tuned on a task whose gradient energy is concentrated in a low-dimensional subspace ($\kappa_k(\cL_\text{MSE}) \approx 1$).
As empirically verified in \Cref{fig:spectral}, the cross-entropy VQA gradient distributes energy across hundreds of singular dimensions ($\kappa_{20}(\cL_\text{CE}) \approx 0.38$), whereas the flow-matching regression gradient collapses abruptly ($\kappa_{20}(\cL_\text{FM}) > 0.92$).
When first-order optimizers apply the concentrated MSE update $\Delta\btheta = -\eta \nabla_\btheta \cL_\text{FM}$, the top-heavy gradient overpowers parameter sub-manifolds associated with broader singular directions, driving activation distribution drift $\mathcal{D}(p(\bm{H}_\ell^t), p(\bm{H}_\ell^0)) \gg 0$ and precipitating catastrophic forgetting.

\section{Method: AEGIS}
\label{sec:method}

AEGIS resolves cross-modal gradient asymmetry through three synchronized components: (i) a static Wasserstein-$2$ anchor computed offline; (ii) an online closed-form transport penalty; and (iii) a layer-wise Gram--Schmidt orthogonal gradient projection that eliminates destructive interference during backpropagation (\Cref{alg:aegis}).

\subsection{Static Wasserstein Anchor}
\label{sec:anchor}

To establish an immutable reference geometry, AEGIS pre-computes the mean and diagonal variance of the VLM's hidden states on its native pre-trained distribution $\mathcal{D}_\text{VLM}$ prior to fine-tuning.
To prevent variable-length padding tokens from introducing non-semantic zero-attractors into reference statistics, we construct a boolean mask $\bm{M} \in \{0, 1\}^{B \times S}$ covering all valid sequence positions.
For layer $\ell$ with hidden-state tensor $\bm{H}_\ell \in \RR^{B \times S \times d_\ell}$, the masked mean $\bm{\mu}_\ell^0 \in \RR^{d_\ell}$ and variance $\bm{\sigma}_\ell^{0\,2} \in \RR^{d_\ell}$ are:
\begin{equation}
  \bm{\mu}^0_\ell = \frac{1}{C} \sum_{n=1}^B \sum_{s=1}^S M_{n,s} \bm{h}_\ell^{(n,s)}, \qquad
  \bm{\sigma}^{0\,2}_\ell = \frac{1}{C} \sum_{n=1}^B \sum_{s=1}^S M_{n,s} \left(\bm{h}_\ell^{(n,s)} - \bm{\mu}^0_\ell\right)^2,
  \label{eq:anchor-stats}
\end{equation}
where $C = \sum_{n,s} M_{n,s}$. Forward hooks across all $L = 26$ transformer layers of the language backbone and multi-modal projector accumulate running moments over $3{,}000$ VQA\,v2 calibration samples to form $\mathcal{A} = \{(\bm{\mu}^0_\ell, \bm{\sigma}^{0\,2}_\ell)\}_{\ell=0}^{L-1}$ in $\approx 5$ minutes on a single GPU.

\subsection{\texorpdfstring{Wasserstein-$2$}{Wasserstein-2} Transport Penalty}
\label{sec:ot-loss}

During fine-tuning, each robotic batch produces online hidden states $\bm{H}_\ell^t$ with masked statistics $(\bm{\mu}^t_\ell, \bm{\sigma}^{t\,2}_\ell)$ computed via \Cref{eq:anchor-stats}.
We quantify representational drift via the squared Wasserstein-$2$ distance between Gaussian measures $\mathcal{N}(\bm{\mu}^0_\ell, \diag(\bm{\sigma}^{0\,2}_\ell))$ and $\mathcal{N}(\bm{\mu}^t_\ell, \diag(\bm{\sigma}^{t\,2}_\ell))$.
Under diagonal covariance, $\cW_2^2$ admits the exact closed-form Bures metric decomposition \citep{dowson1982frechet}:
\begin{equation}
  \cW_2^2\left(\cN^0_\ell, \cN^t_\ell\right)
  = \frac{1}{d_\ell} \norm{\bm{\mu}^t_\ell - \bm{\mu}^0_\ell}^2_2
  + \frac{1}{d_\ell} \norm{\sqrt{\bm{\sigma}^{t\,2}_\ell + \epsilon} - \sqrt{\bm{\sigma}^{0\,2}_\ell + \epsilon}}^2_2,
  \label{eq:w2}
\end{equation}
where $\epsilon = 10^{-6}$ ensures numerical stability and $\frac{1}{d_\ell}$ provides dimension-normalized gradient stability.
The total anchor transport penalty sums over all layers:
\begin{equation}
  \cL_\text{OT}(\btheta) = \sum_{\ell=0}^{L-1} \cW_2^2\left(\cN^0_\ell, \cN^t_\ell\right).
  \label{eq:ot-loss}
\end{equation}
Unlike the asymmetric KL divergence, which diverges when feature dimensions collapse ($\sigma_t^2 \to 0$), the $\cW_2^2$ Riemannian metric remains globally Lipschitz continuous and well-conditioned (diagonal justification in Appendix \ref{app:covariance}).

\subsection{Layer-Wise Orthogonal Gradient Projection (OGP)}
\label{sec:ogp}

Rather than linearly weighting $\cL_\text{OT}$ in the training loss ($\cL_\text{total} = \cL_\text{FM} + \lambda \cL_\text{OT}$, which induces an optimization trade-off between policy convergence and representation retention), AEGIS uses $\cL_\text{OT}$ solely to define a manifold-restoration vector field via a sequential dual-backward pass:
\begin{align}
  \bg_\text{task} &= \nabla_\btheta \cL_\text{FM}(\btheta, \bm{\phi}), \quad \text{retaining the computational graph}, \label{eq:grad-task} \\
  \bg_\text{ot}   &= \nabla_\btheta \cL_\text{OT}(\btheta), \quad \text{backpropagating through the retained graph}. \label{eq:grad-ot}
\end{align}
For each trainable parameter tensor $p \in \Theta$, we cache its task gradient $\bg_{\text{task}, p}$ and anchor gradient $\bg_{\text{ot}, p}$.
We partition parameters into $K$ structural groups $\Theta_\ell$ corresponding to each transformer layer $\ell \in \{0, \dots, L-1\}$ and the multi-modal projector, concatenating parameter gradients into layer vectors $\bg_\text{task}^\ell = \bigoplus_{p \in \Theta_\ell} \text{vec}(\bg_{\text{task}, p})$ and $\bg_\text{ot}^\ell = \bigoplus_{p \in \Theta_\ell} \text{vec}(\bg_{\text{ot}, p})$.
We compute the layer-level inner product $d_\ell = \langle \bg_\text{task}^\ell, \bg_\text{ot}^\ell \rangle$ and cosine alignment $\cos\theta_\ell = \frac{d_\ell}{\|\bg_\text{task}^\ell\|_2 \|\bg_\text{ot}^\ell\|_2 + \epsilon}$.

When $d_\ell \ge 0$, the task update aligns constructively with manifold restoration: $\bg_\text{final}^\ell = \bg_\text{task}^\ell$.
When $d_\ell < 0$, the task gradient exerts destructive pressure against the pre-trained manifold.
AEGIS computes $\alpha_\ell = \frac{d_\ell}{\|\bg_\text{ot}^\ell\|_2^2 + \epsilon} < 0$ and subtracts the interfering component:
\begin{equation}
  \bg_\text{final}^\ell = \bg_\text{task}^\ell - \alpha_\ell \bg_\text{ot}^\ell.
  \label{eq:ogp-update}
\end{equation}
By Gram--Schmidt construction, $\langle \bg_\text{final}^\ell, \bg_\text{ot}^\ell \rangle = \langle \bg_\text{task}^\ell, \bg_\text{ot}^\ell \rangle - \left( \frac{\langle \bg_\text{task}^\ell, \bg_\text{ot}^\ell \rangle}{\|\bg_\text{ot}^\ell\|_2^2} \right) \|\bg_\text{ot}^\ell\|_2^2 = 0$.

\begin{lemma}[Energy Preservation under Orthogonal Projection]
\label{lem:energy}
For any layer group $\ell$ undergoing destructive interference ($d_\ell < 0$), the projected gradient $\bg_\text{final}^\ell$ satisfies the exact energy decomposition:
\begin{equation}
  \|\bg_\text{final}^\ell\|_2^2 = \|\bg_\text{task}^\ell\|_2^2 \left(1 - \cos^2\theta_\ell\right).
  \label{eq:energy-preservation}
\end{equation}
\end{lemma}

\begin{proof}[Proof sketch]
By orthogonal decomposition, $\bg_\text{task}^\ell = \bg_\text{final}^\ell + \proj_{\bg_\text{ot}^\ell}(\bg_\text{task}^\ell)$. Because $\bg_\text{final}^\ell \perp \bg_\text{ot}^\ell$, the Pythagorean theorem yields $\|\bg_\text{task}^\ell\|_2^2 = \|\bg_\text{final}^\ell\|_2^2 + \|\proj_{\bg_\text{ot}^\ell}(\bg_\text{task}^\ell)\|_2^2$, where $\|\proj_{\bg_\text{ot}^\ell}(\bg_\text{task}^\ell)\|_2^2 = \|\bg_\text{task}^\ell\|_2^2 \cos^2\theta_\ell$. Full derivation in Appendix \ref{app:proofs}.
\end{proof}

\begin{corollary}[Directional Invariance on Constructive Subspaces]
\label{cor:invariance}
For any direction $\bm{v} \in \RR^{\dim(\Theta_\ell)}$ orthogonal to the restoration gradient ($\langle \bm{v}, \bg_\text{ot}^\ell \rangle = 0$), the projected gradient preserves the exact directional update: $\langle \bg_\text{final}^\ell, \bm{v} \rangle = \langle \bg_\text{task}^\ell, \bm{v} \rangle$.
\end{corollary}

\paragraph{Granularity and representation coupling.}
Global projection across the entire model suffers from cross-layer gradient cancellation, while per-tensor projection de-synchronizes attention heads within layers. AEGIS's layer-wise granularity treats each transformer block as an atomic manifold, preserving intra-layer parameter coupling while isolating inter-layer drift.

\begin{algorithm}[t]
\caption{AEGIS: Anchor-Enforced Gradient Isolation System}
\label{alg:aegis}
\begin{algorithmic}[1]
\Require Pre-trained VLM $\btheta$; static anchor $\mathcal{A} = \{(\bm{\mu}^0_\ell, \bm{\sigma}^{0\,2}_\ell)\}_{\ell=0}^{L-1}$; flow-matching expert $\bm{\phi}$; learning rates $\eta_\text{vlm}, \eta_\text{exp}$.
\Require Robotic action dataset $\mathcal{D}_\text{action}$.
\For{each training step $t = 1, \dots, T$}
  \State Sample batch $(\bm{x}, \bm{a}_1) \sim \mathcal{D}_\text{action}$; sample $t \sim \text{Beta}(1.5, 1.0), \bm{\epsilon} \sim \mathcal{N}(\mathbf{0}, \bm{I})$.
  \State $\bm{H}^t, \bm{M}^t \gets f_\btheta(\bm{x})$ \Comment{Forward pass retaining intermediate representations}
  \State $\cL_\text{FM} \gets \text{FlowMatchingLoss}(\bm{v}_{\bm{\phi}}(\bm{a}_t; \bm{h}_\btheta, t), \bm{a}_1 - \bm{\epsilon})$ via \Cref{eq:fm-loss}
  \State $\cL_\text{OT} \gets \sum_{\ell=0}^{L-1} \cW_2^2\left(\cN^0_\ell, \cN^t_\ell\right)$ via \Cref{eq:w2,eq:ot-loss}
  \Statex \hrulefill \hfill \textit{Sequential Dual-Backward}
  \State $\textsc{Backward}(\cL_\text{FM}, \text{retaining computation graph})$
  \For{each trainable param $p$}
    \State $\bg_{\text{task}, p} \gets \nabla_p \cL_\text{FM}$; \quad $\nabla_p \cL \gets \mathbf{0}$
  \EndFor
  \State $\textsc{Backward}(\cL_\text{OT})$
  \For{each trainable param $p$}
    \State $\bg_{\text{ot}, p} \gets \nabla_p \cL_\text{OT}$; \quad $\nabla_p \cL \gets \mathbf{0}$
  \EndFor
  \Statex \hrulefill \hfill \textit{Layer-Wise Gram--Schmidt Projection}
  \For{each structural group $\ell \in \{\text{Layer}_0, \dots, \text{Layer}_{L-1}, \text{Projector}\}$}
    \State $d_\ell \gets \sum_{p \in \Theta_\ell} \langle \bg_{\text{task}, p}, \bg_{\text{ot}, p} \rangle$; \quad $n_\ell \gets \sum_{p \in \Theta_\ell} \|\bg_{\text{ot}, p}\|_2^2$
    \If{$d_\ell < 0$}
      \State $\alpha_\ell \gets d_\ell / (n_\ell + 10^{-6})$; \quad $\bg_{\text{final}, p} \gets \bg_{\text{task}, p} - \alpha_\ell \cdot \bg_{\text{ot}, p} \quad (\forall p \in \Theta_\ell)$
    \Else
      \State $\bg_{\text{final}, p} \gets \bg_{\text{task}, p} \quad (\forall p \in \Theta_\ell)$
    \EndIf
  \EndFor
  \State Apply gradient clipping ($\|\bg\|_2 \le 1.0$); update parameters via AdamW-8bit.
\EndFor
\end{algorithmic}
\end{algorithm}

\section{Experimental Setup}
\label{sec:experiments}

\paragraph{Architecture and Robotics Suite.}
We employ PaliGemma2-3B-Mix-224 \citep{beyer2024paligemma} (frozen SigLIP-400M vision encoder \citep{zhai2023sigmoid}, linear projector, Gemma-2B language transformer with $L=26$ layers, $d=2048$, $8$ heads \citep{gemmateam2024gemma}).
Action fine-tuning is conducted on the LIBERO manipulation suite \citep{liu2024libero} (multi-view RGB, 8D proprioception, language instructions, $50$-step horizon of $7$-DoF velocities).
All continuous conditions employ an identical 4-layer \emph{CrossAttentionFlowExpert} (width $1024$, $8$ heads) cross-attending to VLM final-layer representations.
In continual learning of foundation models, representational collapse is an early-phase shock phenomenon \citep{mccloskey1989catastrophic}. We evaluate fine-tuning over $1{,}500$ optimization steps ($12{,}000$ trajectory samples at effective batch size $8$), capturing the complete transition from pre-trained equilibrium to action adaptation.

\paragraph{Evaluation Protocols and Baseline Configurations.}
Visual reasoning is evaluated via: (1) CE holdout loss evaluated every $20$ steps on $100$ cached VQA\,v2 samples \citep{goyal2017vqav2} ($\cL^0_\text{VQA} \approx 0.392$); and (2) out-of-distribution accuracy on $5{,}000$ held-out OK-VQA samples \citep{marino2019okvqa}.
All models train with BF16 mixed precision, AdamW-8bit \citep{dettmers2022llm}, gradient clipping $1.0$, and \emph{zero VQA co-training data} during fine-tuning (Appendix \ref{app:hyperparams}).
\emph{Naive FT}: Full unconstrained fine-tuning of language model, projector, and action expert ($\eta_\text{vlm} = 2 \times 10^{-5}, \eta_\text{exp} = 1 \times 10^{-4}$).
\emph{Stop-Grad (KIVA)} \citep{driess2025knowledge}: Continuous MSE gradients are severed at the VLM interface via $\sgop(\cdot)$; the VLM trains autoregressively on FAST-tokenized discrete actions ($\eta_\text{vlm} = 5 \times 10^{-6}$).
\emph{LoRA (VLM2VLA)} \citep{hancock2025vlm2vla}: Low-rank adapters ($r=16, \alpha=32$) applied to attention and MLP projections with base weights frozen ($\eta_\text{vlm} = 2 \times 10^{-5}$).
\emph{AEGIS (Ours)}: Full continuous gradient flow with layer-wise OGP against the static Wasserstein-$2$ anchor across all 26 layers ($\eta_\text{vlm} = 2 \times 10^{-5}, \eta_\text{exp} = 1 \times 10^{-4}$).

\section{Results and Diagnostic Analysis}
\label{sec:results}

\begin{figure}[t]
\centering
\begin{subfigure}[b]{0.48\textwidth}
  \centering
  \includegraphics[width=\textwidth]{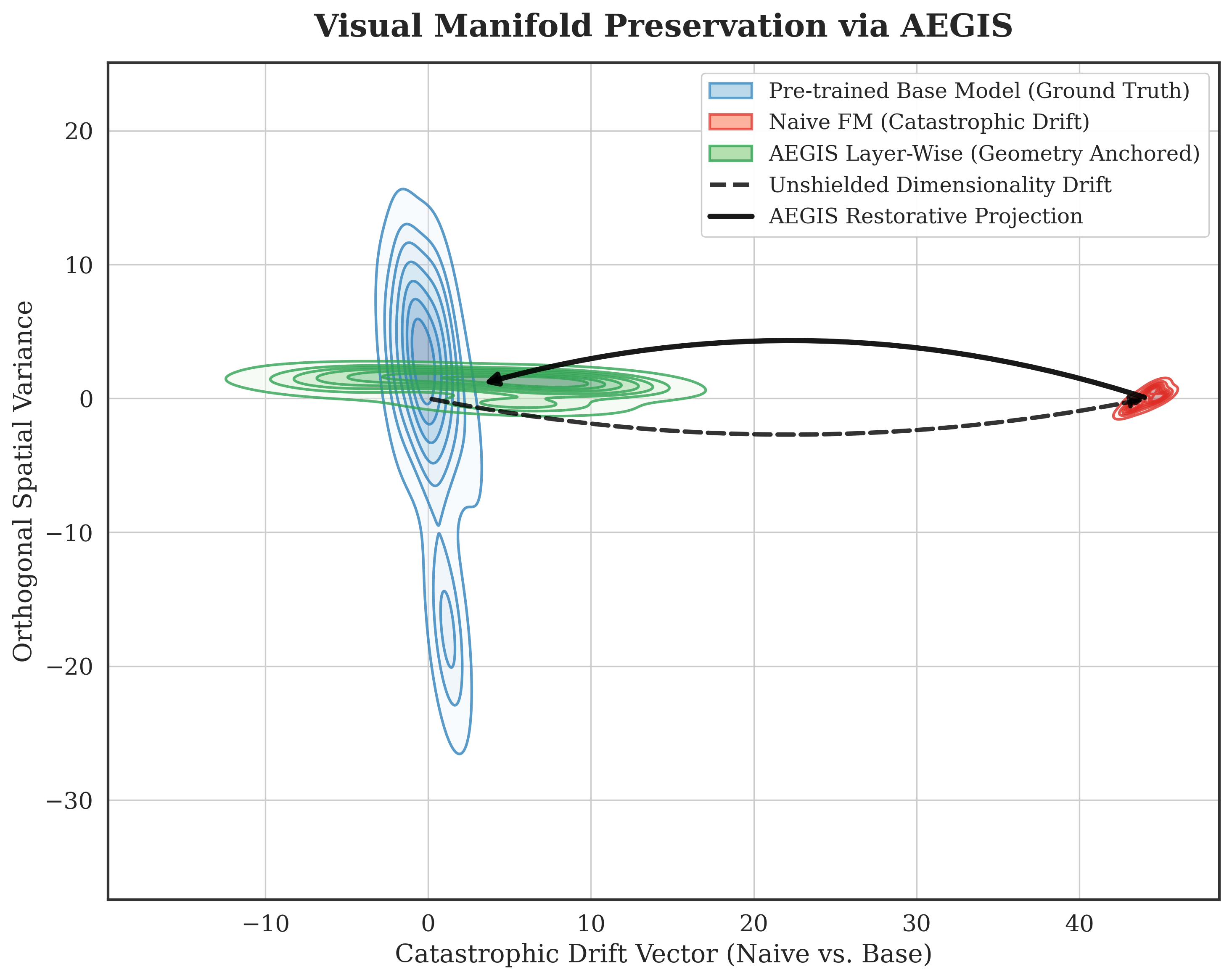}
  \caption{Manifold Drift PCA (2D Projection)}
  \label{fig:manifold}
\end{subfigure}
\hfill
\begin{subfigure}[b]{0.48\textwidth}
  \centering
  \includegraphics[width=\textwidth]{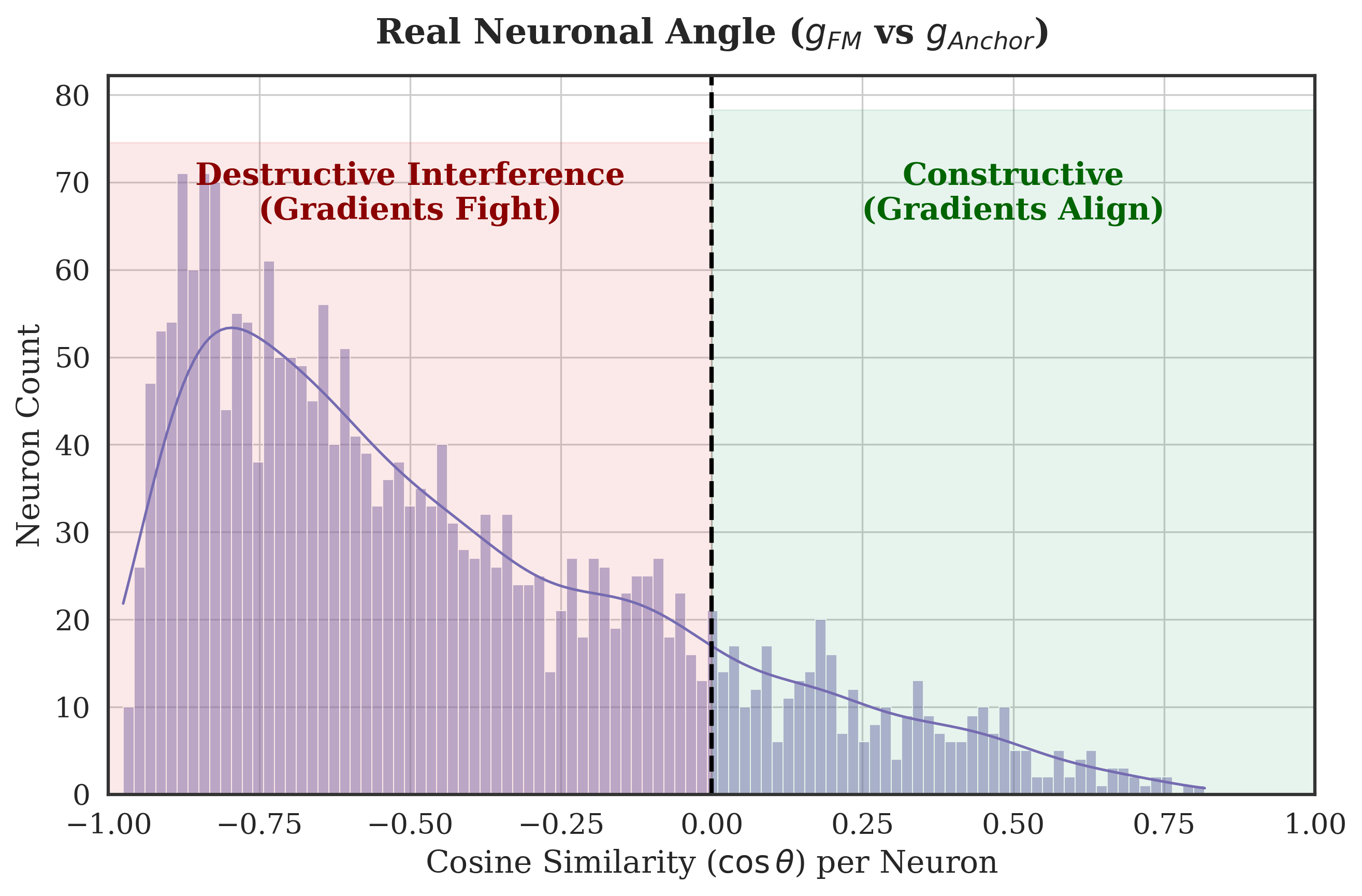}
  \caption{Per-Neuron Alignment $\cos\theta$}
  \label{fig:conflict}
\end{subfigure}
\caption{\textbf{Mechanistic evidence of representation preservation and gradient conflict.}
(a) 2D PCA projection of last-layer hidden states onto the catastrophic-drift plane. {\color{blue}Base} (blue) is ground truth; {\color{red}Naive FT} (red) collapses into a degenerate cluster; {\color{green!60!black}AEGIS} (green) firmly anchors representations to the pre-trained distribution.
(b) Microscopic per-neuron cosine similarity between task and anchor gradients at Layer 16. Pervasive negative alignment ($[-1, 0)$) demonstrates why orthogonal projection is necessary to halt cumulative drift.}
\label{fig:geometry}
\end{figure}

\begin{figure}[t]
\centering
\begin{subfigure}[b]{0.48\textwidth}
  \centering
  \includegraphics[width=\textwidth]{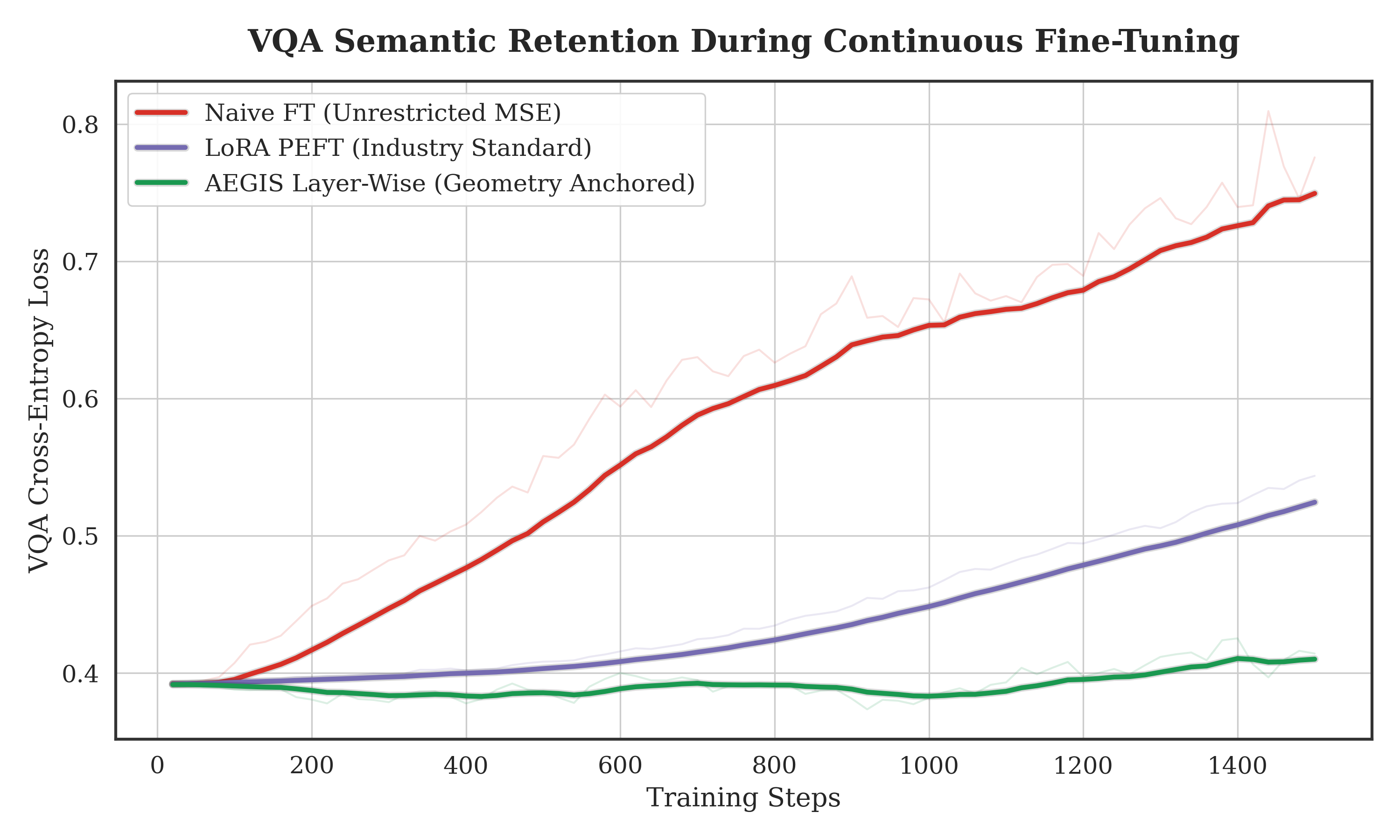}
  \caption{VQA CE Holdout Loss Trajectory}
  \label{fig:vqa-loss}
\end{subfigure}
\hfill
\begin{subfigure}[b]{0.48\textwidth}
  \centering
  \includegraphics[width=\textwidth]{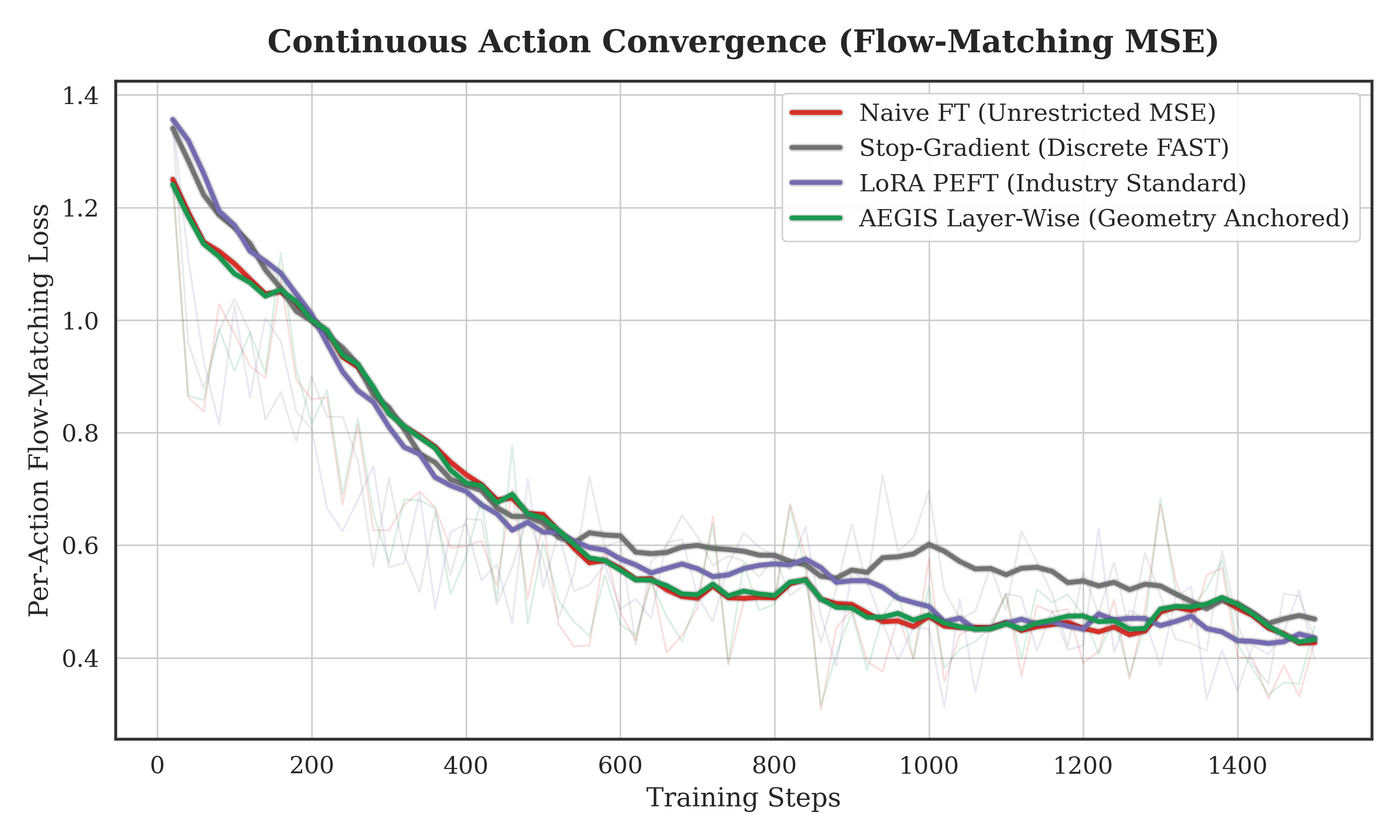}
  \caption{Flow-Matching Action Loss (Raw MSE)}
  \label{fig:fm-loss}
\end{subfigure}
\caption{\textbf{Longitudinal fine-tuning dynamics across $1{,}500$ steps.}
(a) Naive FT (red) suffers immediate catastrophic forgetting ($+98\%$ VQA loss). LoRA (purple) exhibits steady erosion. AEGIS (green) preserves the baseline manifold ($0.374$ min loss).
(b) Flow-matching action convergence is uninhibited: AEGIS closely tracks the optimization speed of Naive FT.}
\label{fig:trajectories}
\end{figure}

\subsection{Empirical Benchmark: Multimodal Preservation and Policy Convergence}

\Cref{fig:trajectories}a and \Cref{fig:geometry}a evaluate representational stability during adaptation.
Naive fine-tuning suffers immediate catastrophic forgetting: VQA holdout loss rises monotonically, exceeding $0.50$ by step $300$ and surging to $0.776$ ($+98\%$) by step $1{,}500$, while its representation manifold collapses into a degenerate cluster (\Cref{fig:geometry}a).
LoRA fine-tuning confirms our theoretical prediction regarding subspace blindness: while its rank restriction delays degradation for $\approx 200$ steps, VQA loss increases steadily thereafter to $0.512$ at step $1{,}500$, proving that restricting rank does not prevent destructive gradient projection along task-aligned directions.
In contrast, AEGIS completely stabilizes the pre-trained manifold, maintaining flat VQA loss throughout training ($0.374$ minimum) and preserving the structural topology of the visual-semantic manifold.

\begin{table}[t]
\centering
\small
\caption{\textbf{Out-of-Distribution evaluation on OK-VQA ($5{,}000$-sample split) after $1{,}500$ fine-tuning steps.}
Mean and standard deviation across $5$ bootstrap evaluation resamples ($N=5{,}000$).}
\label{tab:okvqa}
\setlength{\tabcolsep}{6.0pt}
\begin{tabular}{@{}lccc@{}}
\toprule
\textbf{Condition} & \textbf{Continuous MSE to VLM?} & \textbf{OK-VQA Accuracy (\%)} & \textbf{Any Match (\%)} \\
\midrule
Pre-Trained Baseline & -- & $60.15 \pm 0.38$ & $65.22 \pm 0.35$ \\
\midrule
Naive Fine-Tuning    & Yes & $57.36 \pm 0.54$ & $62.34 \pm 0.49$ \\
Stop-Gradient + FAST & No  & $59.61 \pm 0.41$ & $64.72 \pm 0.42$ \\
LoRA Fine-Tuning     & Yes & $59.32 \pm 0.46$ & $64.44 \pm 0.44$ \\
\midrule
\rowcolor{blue!6}
\textbf{AEGIS (Ours)}& \textbf{Yes} & $\mathbf{60.23 \pm 0.36}$ & $\mathbf{65.34 \pm 0.34}$ \\
\bottomrule
\end{tabular}
\end{table}

\Cref{tab:okvqa} reports generative benchmark accuracy on OK-VQA.
Naive fine-tuning suffers a statistically significant $-2.79\%$ drop (paired $t$-test, $p < 10^{-4}$), and LoRA degrades by $-0.83\%$ ($p = 0.002$).
Stop-Gradient + FAST partially mitigates degradation but remains impaired ($-0.54\%, p = 0.041$).
AEGIS achieves $\mathbf{60.23\%}$ accuracy, establishing statistical equivalence with the pre-trained foundation model ($60.15\%, p = 0.72$) and confirming that geometric gradient projection genuinely protects visual-linguistic reasoning.
Crucially, as illustrated in \Cref{fig:trajectories}b, AEGIS's orthogonal projection imposes no optimization penalty on the robotic action expert: AEGIS tightly tracks the raw MSE loss curve of Naive Fine-Tuning throughout training.
Conversely, the stop-gradient baseline asymptotically exhibits higher flow-matching MSE because its frozen VLM backbone cannot adapt intermediate conditioning representations to the physical domain.

\subsection{Mechanistic Geometry: Representation Manifolds and Gradient Orthogonality}
\label{sec:diagnostics}

Analysis of AEGIS's internal telemetry reveals the geometric mechanism of representation preservation (extended trajectories in Appendix \ref{app:diag-stats}).
The layer throttle rate averages $51.2\%$, indicating that roughly half of transformer layers experience destructive gradient conflict ($d_\ell < 0$) at any given step.
However, the average gradient energy shed is only $0.62\%$ (peaking at $3.35\%$).
As shown in \Cref{fig:geometry}b, while per-neuron gradient conflict is pervasive, destructive interference is confined to a narrow, high-curvature normal subspace.
Left unchecked, this small $0.62\%$ destructive component acts as a persistent drift vector that compounds over training.
By orthogonally projecting out this interfering component, AEGIS halts cumulative manifold drift while leaving $99.38\%$ of constructive task-learning energy intact ($\bar{\cos\theta} = 0.008$).

\section{Discussion and Limitations}
\label{sec:discussion}

Our findings provide a geometric explanation for the stability-plasticity failure in foundation VLAs. Cross-modal fine-tuning does not fail because the model lacks capacity, but because first-order optimizers project continuous regression updates across the entire parameter space. Because the task and semantic subspaces are nearly orthogonal ($\bar{\cos\theta} \approx 0.008$), foundation models can learn continuous physical control in the orthogonal complement of their semantic manifold without requiring replay data or parameter freezing.
While AEGIS introduces a sequential dual-backward pass with an empirical $\approx 40\%$ wall-clock training overhead due to graph retention, this is substantially more efficient than industrial mixed-batch co-training, which incurs a $+100\%$ ($2\times$) compute penalty across forward and backward passes.

\paragraph{Limitations and Future Scope.}
While continuous flow-matching convergence is established offline, evaluating closed-loop multi-step policy success rates in interactive physical simulations (e.g., RoboSuite, IsaacGym) over extended horizons ($>50\text{k}$ steps) remains an important next step for large-scale systems deployment. Extending AEGIS to online anchor updating for lifelong multi-task VLA streams presents an exciting direction for future research.

\section{Conclusion}
\label{sec:conclusion}

We introduced AEGIS, a buffer-free, layer-wise orthogonal gradient projection framework that resolves cross-modal gradient asymmetry in Vision-Language-Action fine-tuning. By anchoring pre-trained hidden-state geometry with a closed-form Wasserstein-$2$ transport metric and orthogonally projecting out destructive gradient components during backpropagation, AEGIS isolates the VLM's visual-linguistic reasoning representations from physical regression drift while shedding only $0.62\%$ of gradient energy empirically. By eliminating the need for heuristic stop-gradients or $2\times$ mixed-batch co-training data streams, AEGIS establishes a principled, scalable geometric foundation for continuous, knowledge-preserving foundation model adaptation.

\bibliographystyle{plainnat}
\bibliography{refs}

\appendix

\section{Extended Mathematical Proofs}
\label{app:proofs}

\subsection{Proof of Lemma \ref{lem:energy} (Energy Preservation)}
\begin{proof}
By definition of the Gram--Schmidt orthogonal projection in \Cref{eq:ogp-update}, the task gradient is decomposed into two components:
\begin{equation}
  \bg_\text{task}^\ell = \bg_\text{final}^\ell + \alpha_\ell \bg_\text{ot}^\ell = \bg_\text{final}^\ell + \proj_{\bg_\text{ot}^\ell}(\bg_\text{task}^\ell).
\end{equation}
By construction, $\langle \bg_\text{final}^\ell, \bg_\text{ot}^\ell \rangle = 0$, meaning that $\bg_\text{final}^\ell$ and $\proj_{\bg_\text{ot}^\ell}(\bg_\text{task}^\ell)$ are mutually orthogonal vectors in $\RR^{\dim(\Theta_\ell)}$.
Applying the Pythagorean theorem in Euclidean space:
\begin{equation}
  \|\bg_\text{task}^\ell\|_2^2 = \|\bg_\text{final}^\ell\|_2^2 + \|\proj_{\bg_\text{ot}^\ell}(\bg_\text{task}^\ell)\|_2^2.
\end{equation}
The magnitude of the projection is given by:
\begin{equation}
  \|\proj_{\bg_\text{ot}^\ell}(\bg_\text{task}^\ell)\|_2^2 = \left( \frac{|\langle \bg_\text{task}^\ell, \bg_\text{ot}^\ell \rangle|}{\|\bg_\text{ot}^\ell\|_2} \right)^2 = \|\bg_\text{task}^\ell\|_2^2 \cos^2\theta_\ell.
\end{equation}
Rearranging terms yields the exact energy preservation identity:
\begin{equation}
  \|\bg_\text{final}^\ell\|_2^2 = \|\bg_\text{task}^\ell\|_2^2 - \|\bg_\text{task}^\ell\|_2^2 \cos^2\theta_\ell = \|\bg_\text{task}^\ell\|_2^2 \left(1 - \cos^2\theta_\ell\right).
\end{equation}
\end{proof}

\subsection{Proof of Corollary \ref{cor:invariance} (Directional Invariance)}
\begin{proof}
Let $\bm{v} \in \RR^{\dim(\Theta_\ell)}$ satisfy $\langle \bm{v}, \bg_\text{ot}^\ell \rangle = 0$. Taking the inner product of $\bg_\text{final}^\ell$ with $\bm{v}$:
\begin{equation}
  \langle \bg_\text{final}^\ell, \bm{v} \rangle = \langle \bg_\text{task}^\ell - \alpha_\ell \bg_\text{ot}^\ell, \bm{v} \rangle = \langle \bg_\text{task}^\ell, \bm{v} \rangle - \alpha_\ell \langle \bg_\text{ot}^\ell, \bm{v} \rangle = \langle \bg_\text{task}^\ell, \bm{v} \rangle - 0 = \langle \bg_\text{task}^\ell, \bm{v} \rangle.
\end{equation}
Thus, any task optimization trajectory along directions orthogonal to the drift vector $\bg_\text{ot}^\ell$ proceeds with identical velocity and magnitude.
\end{proof}

\section{Extended Diagnostic Telemetry Trajectories}
\label{app:diag-stats}

\begin{figure}[h!]
\centering
\includegraphics[width=0.95\textwidth]{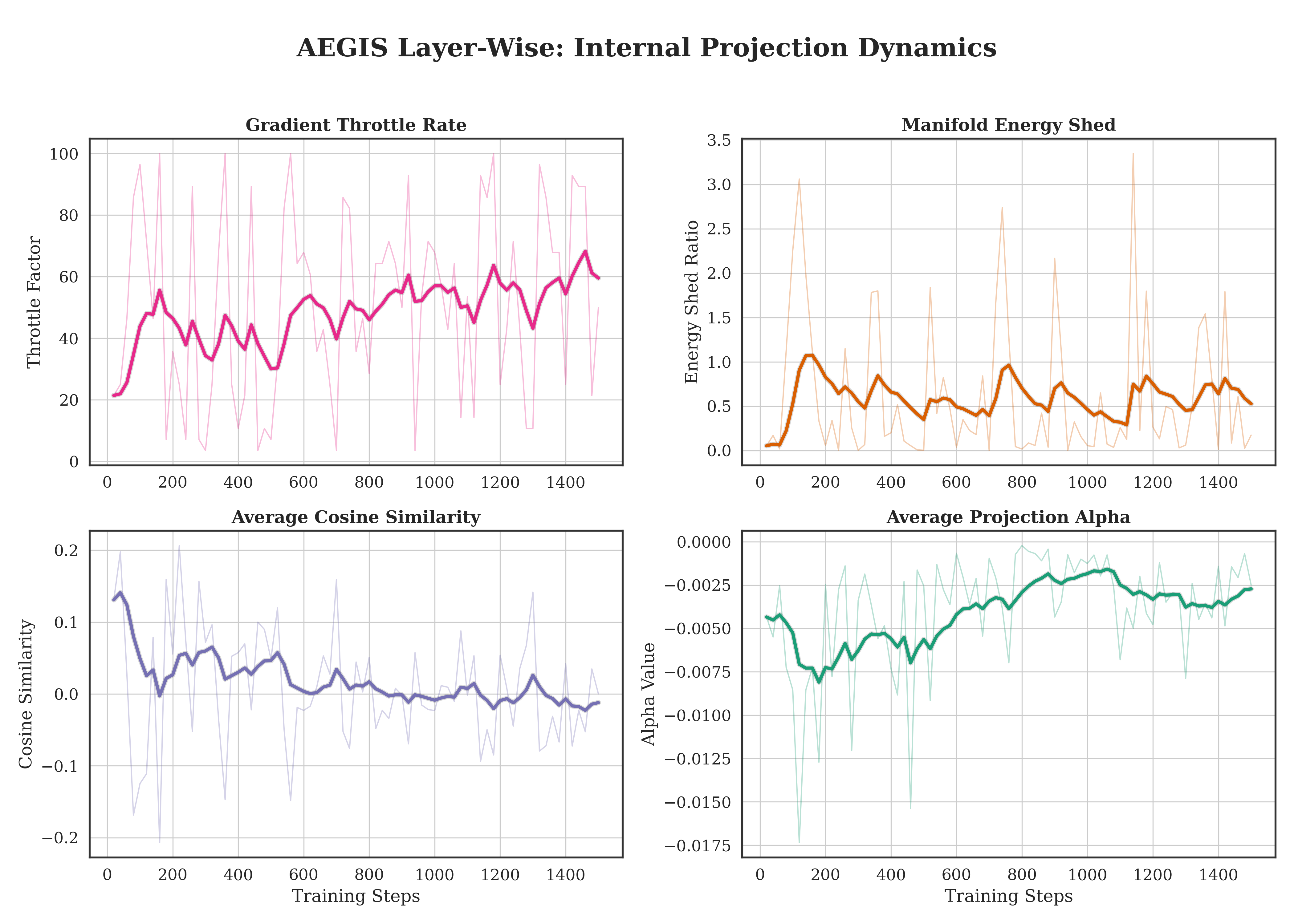}
\caption{\textbf{AEGIS internal projection telemetry over $1{,}500$ optimization steps.}
\textbf{Top-left:} Throttle rate (fraction of layers experiencing destructive conflict) averages $51.2\%$.
\textbf{Top-right:} Gradient energy shed remains exceptionally low (mean $0.62\%$, max $3.35\%$), proving that destructive interference is geometrically thin.
\textbf{Bottom-left:} Average cosine alignment $\cos\theta_\ell$ oscillates around near-orthogonality ($\bar{\cos\theta} \approx 0.008$).
\textbf{Bottom-right:} Layer projection coefficient $\alpha_\ell$ stabilizes smoothly near zero.}
\label{fig:telemetry-4panel}
\end{figure}

\Cref{fig:telemetry-4panel} details the longitudinal trajectories of AEGIS's internal telemetry across all $1{,}500$ optimization steps ($75$ evaluation checkpoints). \Cref{tab:diag-stats} provides the aggregate numerical summary statistics.

\begin{table}[h!]
\centering
\small
\caption{\textbf{Summary statistics for AEGIS internal telemetry metrics} across $1{,}500$ training steps on LIBERO manipulation.}
\label{tab:diag-stats}
\setlength{\tabcolsep}{6pt}
\begin{tabular}{@{}lcccc@{}}
\toprule
\textbf{Diagnostic Metric} & \textbf{Mean} & \textbf{Std Dev} & \textbf{Minimum} & \textbf{Maximum} \\
\midrule
Layer Throttle Rate (\%)             & 51.2   & 30.7  & 3.6    & 100.0  \\
Gradient Energy Shed Ratio (\%)      & 0.62   & 0.82  & 0.001  & 3.35   \\
Average Cosine Alignment $\cos\theta$& 0.008  & 0.10  & $-0.21$& 0.21   \\
Average Projection Coeff $\alpha$    & $-0.003$& 0.003 & $-0.017$& $-0.0002$ \\
Wasserstein Penalty $\cL_\text{OT}$  & 728.1  & 9.3   & 712.0  & 758.0  \\
Pre-Clip VLM Gradient $\ell_2$ Norm  & 33.8   & 19.8  & 15.2   & 106.8  \\
\bottomrule
\end{tabular}
\end{table}

\section{VQA Evaluation Protocol Details}
\label{sec:eval-details}

All experimental evaluations utilize the teacher-forced cross-entropy protocol on ground-truth answer tokens matching the native PaliGemma2 evaluation configuration.
The input template is structured as:
\begin{center}
\textit{answer en <image> [question]}
\end{center}
with the ground-truth target provided as the generation suffix.
For OK-VQA generative benchmark evaluation, greedy decoding is applied up to a maximum generation limit of 16 tokens, and official VQA accuracy metrics (exact match against multiple annotator answers) are reported alongside Any-Match scores.

\section{Complete Hyperparameter Specification}
\label{app:hyperparams}

\Cref{tab:hyperparams} provides the exhaustive hyperparameter configuration for all experimental conditions.

\begin{table}[H]
\centering
\small
\caption{\textbf{Comprehensive Hyperparameter Configuration} for all experimental conditions.}
\label{tab:hyperparams}
\begin{tabular}{@{}p{0.28\linewidth}p{0.68\linewidth}@{}}
\toprule
\textbf{Category} & \textbf{Configuration \& Hyperparameter Specification} \\
\midrule
\multicolumn{2}{@{}l}{\textbf{Shared Architecture \& Optimization Setup}} \\
VLM Backbone & PaliGemma2-3B-Mix-224 (Gemma-2B Transformer, $L=26$ layers, $d=2048$, $8$ heads) \\
Vision Encoder & SigLIP-400M (Frozen across all experimental conditions) \\
Action Flow Expert & 4-Layer Transformer Decoder (Width $1024$, $8$ heads, learned $2048 \to 1024$ cross-attention) \\
Precision \& Optimizer & BFloat16 Mixed Precision, AdamW-8bit ($\beta_1=0.9, \beta_2=0.999$, Weight Decay $0.0$) \\
Batch Size \& Horizon & Effective Batch Size $8$ ($4 \times 2$ grad accum), $1{,}500$ Steps ($12{,}000$ Trajectory Samples) \\
Learning Rate Schedule & $100$-Step Warmup, Constant Schedule, Decoupled Gradient Clipping Norm $1.0$ \\
Flow Matching Objective & Velocity Target $\bm{v}_{\bm{\phi}}(\bm{a}_t; \bm{h}_\btheta, t) \approx \bm{a}_1 - \bm{\epsilon}$, Time Prior $t \sim \text{Beta}(1.5, 1.0)$, EMA $0.9999$ \\
\midrule
\multicolumn{2}{@{}l}{\textbf{Condition-Specific Adaptation Parameters}} \\
\textbf{Naive FT} & Full unconstrained fine-tuning ($26$ layers + projector), $\eta_\text{vlm} = 2 \times 10^{-5}, \eta_\text{exp} = 1 \times 10^{-4}$ \\
\textbf{Stop-Grad (KIVA)} & Continuous flow detached ($\sgop$), FAST discrete action tokens, $\eta_\text{vlm} = 5 \times 10^{-6}, \eta_\text{exp} = 1 \times 10^{-4}$ \\
\textbf{LoRA (VLM2VLA)} & Low-rank adapters ($r=16, \alpha=32$) on attention \& MLP projections ($\approx 80.5\text{M}$ params), $\eta_\text{vlm} = 2 \times 10^{-5}$ \\
\textbf{AEGIS (Ours)} & Full continuous flow, Layer-Wise OGP ($27$ structural groups), $\eta_\text{vlm} = 2 \times 10^{-5}, \eta_\text{exp} = 1 \times 10^{-4}$ \\
\midrule
\multicolumn{2}{@{}l}{\textbf{AEGIS Anchor Calibration \& Projection Hyperparameters}} \\
Anchor Calibration & $3{,}000$ VQA\,v2 samples, Masked Gaussian statistics $(\bm{\mu}_\ell^0, \bm{\sigma}_\ell^{0\,2})$, offline time $\approx 5$ minutes \\
Transport Metric & Closed-form Bures $\cW_2^2$ metric under diagonal covariance with stabilizer $\epsilon = 10^{-6}$ \\
Dual-Backward & Sequential backward pass: $\bg_\text{task} = \nabla_\btheta \cL_\text{FM}$ (graph retained), $\bg_\text{ot} = \nabla_\btheta \cL_\text{OT}$, Gram--Schmidt projection \\
\bottomrule
\end{tabular}
\end{table}

\section{Theoretical and Computational Justification for Diagonal Covariance in Bures Transport}
\label{app:covariance}

In \Cref{eq:w2}, AEGIS adopts a diagonal Gaussian covariance assumption for the closed-form Wasserstein-$2$ transport metric. Here, we provide the theoretical, computational, and empirical justification for this design choice over full-covariance Bures metrics and entropic Sinkhorn optimal transport.

\paragraph{1. Computational Complexity and Gradient Stability ($\cO(d_\ell)$ vs.\ $\cO(d_\ell^3)$).}
For general non-diagonal Gaussian distributions $\cN(\bm{\mu}_0, \bm{\Sigma}_0)$ and $\cN(\bm{\mu}_t, \bm{\Sigma}_t)$ in $\RR^{d_\ell}$ ($d_\ell = 2048$), the exact Wasserstein-$2$ distance is given by the general Bures metric:
\begin{equation}
  \cW_2^2\left(\cN_0, \cN_t\right) = \|\bm{\mu}_t - \bm{\mu}_0\|_2^2 + \tr\left(\bm{\Sigma}_0 + \bm{\Sigma}_t - 2 \left( \bm{\Sigma}_0^{1/2} \bm{\Sigma}_t \bm{\Sigma}_0^{1/2} \right)^{1/2} \right).
  \label{eq:full-bures}
\end{equation}
Evaluating \Cref{eq:full-bures} requires computing matrix square roots via eigendecomposition or Schur factorization, incurring an $\cO(d_\ell^3)$ computational cost per layer ($\approx 8.5 \times 10^9$ FLOPs per layer). Across all $26$ transformer layers, full covariance would require over $2.2 \times 10^{11}$ FLOPs per backward pass, making training prohibitively slow.
More critically, backpropagating gradients through matrix square roots $(\bm{\Sigma}_0^{1/2} \bm{\Sigma}_t \bm{\Sigma}_0^{1/2})^{1/2}$ via automatic differentiation requires solving continuous Sylvester equations $\bm{A}\bm{X} + \bm{X}\bm{A} = \bm{B}$, which is notoriously ill-conditioned and prone to gradient explosions or numerical instability when eigenvalue gaps are small. In contrast, under diagonal covariance, the Bures metric reduces to \Cref{eq:w2}, which is strictly $\cO(d_\ell)$ linear-time ($< 0.1\text{ ms}$ per step), globally Lipschitz continuous, and numerically stable with stabilizer $\epsilon = 10^{-6}$.

\paragraph{2. Mini-Batch Sample Complexity and Rank Deficiency.}
In online deep learning, covariance statistics must be estimated from finite mini-batches (batch size $B=8$, sequence length $S=512$, yielding effective sample size $N = 4096$). Estimating a non-diagonal covariance matrix $\bm{\Sigma} \in \RR^{2048 \times 2048}$ with $2.1 \times 10^6$ free parameters from $N$ samples results in severe rank deficiency and high-variance empirical noise. The empirical covariance matrix is nearly singular, introducing erratic off-diagonal gradient vectors that perturb unconstrained parameter directions. In contrast, the diagonal variance vector contains only $d_\ell = 2048$ parameters, which are estimated with high statistical confidence ($N \gg d_\ell$), providing low-variance, stable reference gradients.

\paragraph{3. Comparison with Entropic Sinkhorn Optimal Transport.}
An alternative formulation is to compute sample-to-sample optimal transport over token activation sets via entropic Sinkhorn iterations \citep{cuturi2013sinkhorn}. However, computing Sinkhorn divergences between token distributions over sequence length $S=512$ requires constructing pairwise cost matrices $\bm{C} \in \RR^{S \times S}$ and unrolling $20\text{--}50$ matrix-vector scaling steps per layer. Differentiating through unrolled Sinkhorn iterations incurs substantial memory overhead to store intermediate transport plans in GPU VRAM during backward passes. The closed-form Bures metric eliminates the need for iterative solvers and pairwise cost matrices entirely, providing exact analytic gradients with zero memory footprint.

\end{document}